\documentclass[sigconf,nonacm]{acmart}
\usepackage{graphicx}
\usepackage{hyperref}
\usepackage{listings}
\usepackage{booktabs}
\usepackage{multirow}
\usepackage{amsmath}
\usepackage[most]{tcolorbox}
\usepackage{xcolor}
\usepackage{booktabs} 
\usepackage{amsmath}
\usepackage{amsthm}
\usepackage{multirow,tabularx,ragged2e}
\usepackage{adjustbox}
\usepackage{tabularx}
\usepackage{float}
\usepackage{algorithm}
\usepackage{algpseudocode}
\usepackage{makecell}
\usepackage{tcolorbox} 
\usepackage{xcolor}
\usepackage{lipsum}
\usepackage{subcaption}
\usepackage{enumitem}
\usepackage{booktabs}

\newcolumntype{Y}{>{\raggedright\arraybackslash}X}

\newtcolorbox{highlightbox}{
  colback=yellow!20,   
  colframe=black,      
  boxrule=0.5pt,       
  arc=0pt,             
  left=5pt, right=5pt, 
  top=5pt, bottom=5pt  
}

\begin{document}

\title{SSPE: Robust Sequential Staypoint Extraction in Degraded Trajectory Data}
\author{Yueyang Liu}
\orcid{0000-0001-5894-8740}
\affiliation{%
  \institution{Emory University, USA}
  \city{}
  \state{}
  \country{}
}
\email{yueyang.liu@emory.edu}

\author{Lance Kennedy}
\orcid{0009-0004-6815-2219}
\affiliation{%
  \institution{Emory University, USA}
  \city{}
  \state{}
  \country{}
}
\email{lance.kennedy@emory.edu}

\author{Andreas Z{\"u}fle}
\orcid{0000-0001-7001-4123}
\affiliation{%
  \institution{Emory University, USA}
  \city{}
  \state{}
  \country{}
}
\email{azufle@emory.edu}
\renewcommand{\shortauthors}{Liu, et al.}
\begin{abstract}
    Extracting semantically meaningful staypoints from raw trajectory streams is a foundational prerequisite for mobility analytics. However, real-world trajectory data is frequently degraded by severe spatial noise and protracted temporal dropouts. Classic threshold-based extraction methods, such as the widely adopted sliding-window algorithm, are highly sensitive to these anomalies, routinely fracturing continuous stationary episodes into disjointed fragments. To overcome this limitation, we propose the \underline{\textbf{S}}equential \underline{\textbf{S}}tay\underline{\textbf{p}}oint \underline{\textbf{E}}xtraction (SSPE) algorithm. SSPE employs a resilient, state-driven architecture that replaces static spatial anchors with a dynamic centroid and introduces an $n$-step lookahead heuristic to structurally tolerate transient coordinate drift. Concurrently, it utilizes temporal bridging mechanisms to prevent artificial fragmentation during prolonged signal loss. Since optimal extraction parameters vary significantly across datasets, we further introduce a fully unsupervised, consensus-based parameter estimation framework. This framework derives baseline statistical scales directly from the raw trajectory data and identifies the optimal parameter configuration by maximizing structural agreement, thereby eliminating the reliance on annotated ground truth. 
We evaluate SSPE on the ExtraSensory dataset, a real-world mobility corpus with manually verified ground-truth staypoints, against the sliding-window implementation of Trackintel, one of the most widely adopted extraction baselines in the field.
Empirical results demonstrate that SSPE achieves superior accuracy, yielding higher $F_1$-scores while simultaneously improving computational efficiency over this standard.

\end{abstract}

\maketitle

\section{Introduction}

Real-world trajectory data is frequently degraded by severe spatial noise and protracted temporal occlusions. Foundational extraction algorithms, such as the classic sliding-window method~\cite{li2008mining}, rely on rigid, local spatiotemporal thresholds. This reliance makes them exceptionally brittle when applied to sparse or noisy trajectory streams. 
%
Spatial outliers like GPS drift force the detection window to close early, and signal dropouts fracture continuous stays, compromising downstream mobility analytics.

To address these limitations, we propose the Sequential Staypoint Extraction (SSPE) algorithm. 
Rather than relying on the rigid sliding windows of traditional methods or the post-extraction filtering utilized in our prior work~\cite{kennedy2025improvement}, SSPE manages observation uncertainty through a lookahead search and dynamic centroid updating.
Specifically, it replaces the static spatial anchor of traditional methods with a dynamic centroid using an $n$-step lookahead to bridge time gaps without fragmenting the stay. This methodology enables the algorithm to structurally tolerate both extreme coordinate drift and extended signal dropouts without fragmenting continuous stays.

Furthermore, recognizing that the optimal configuration of extraction parameters varies drastically across heterogeneous datasets, we introduce a fully unsupervised parameter estimation framework. 
By extracting baseline noise and temporal scales directly from the raw trajectory data, this framework constructs an empirical search grid and identifies the optimal parameter set by maximizing structural agreement, approximating the performance of supervised tuning entirely without labeled data.

To evaluate the proposed algorithm, we benchmark SSPE against the default sliding-window method implemented in the Trackintel framework~\cite{martin2023trackintel}. As an enhanced adaptation of the foundational trajectory extraction algorithm by Li et al.~\cite{li2008mining}, this baseline provides a standard comparative metric for modern mobility literature. We conduct our empirical analysis exclusively on the ExtraSensory dataset, which is, to the best of our knowledge, the only dataset that contains both high frequency location signals and ground truth staypoint annotations from user-reported activity labels.
To guarantee strict evaluation integrity for this real-world dataset, we manually verified all ground-truth staypoints. 
Experimental results confirm that SSPE yields substantially higher $F_1$-scores than the Trackintel baseline. 
Furthermore, these accuracy improvements are realized alongside a lower runtime than the baseline, demonstrating the algorithm's applicability for large-scale real-world deployment.




\section{Methodology}
This section details our approach to extracting staypoints from noisy, irregularly sampled trajectory data. Our methodology consists of two primary components: the core Sequential Staypoint Extraction Algorithm and an unsupervised framework for its parameter calibration.
\subsection{Sequential Staypoint Extraction Algorithm}
Existing staypoint detection methods like the sliding-window method implemented in commonly-used libraries like Trackintel~\cite{martin2023trackintel} measures each new point against a fixed spatial anchor. Although effective on dense, high-fidelity trajectory datasets, this method degrades considerably under noisy and sparse data.
For example, a single spatial outlier like sensor jitter exceeding the spatial boundary can prematurely terminate the detection window, artificially fragmenting continuous stays into sub-segments that are subsequently discarded.
Furthermore, under sparse sampling, temporal gaps obscure the distinction between continuous dwelling and unobserved transit, causing fragmented or spurious stays~\cite{kennedy2025improvement}.

\subsubsection{Problem definition}

Let $\mathcal{T} = \{p_1, p_2, \dots, p_N\}$ denote a chronologically ordered trajectory of a single agent, where each observation $p = (uid, lat, lon, t)$ consists of the agent identifier, the geographic coordinates, and the timestamp. 
A staypoint is a subsequence of T, contiguous except for skipped transient outliers, in which: (i) each retained observation lies within distance $\delta_{dist}$ of the running centroid of the stay when it is added, (ii) consecutive retained observations are separated by at most $\tau_{max}$, and (iii) the time span from the first to the last retained observation is at least $\tau_{min}$.

Formally, each staypoint is represented as $s = (uid, \mu, t_{\text{first}}, t_{\text{last}})$, where $\mu$ is the representative location and $[t_{\text{first}}, t_{\text{last}}]$ is the dwell interval.

\subsubsection{Algorithm Details}

The SSPE algorithm relies on four core parameters: a spatial distance threshold $\delta_{dist}$, a minimum dwell duration $\tau_{min}$, a maximum temporal dropout tolerance $\tau_{max}$, and a lookahead window $n$. While manual tuning is possible using domain expertise, identifying optimal parameters is generally difficult due to a lack of ground truth for calibration. Therefore, we present a fully unsupervised parameter estimation framework in Section~\ref{sec:estimation} to systematically calibrate these configurations directly from raw data.

Algorithm~\ref{alg:sspe} details the complete extraction procedure. SSPE scans the trajectory $\mathcal{T}$ chronologically using an anchor pointer $i$ and an evaluation pointer $j$. Beginning at an anchor observation $p_i$, the algorithm greedily expands a candidate sequence $\mathcal{C}$ by evaluating each subsequent observation $p_j$ ($j > i$) against strict criteria for temporal continuity and spatial cohesion.
For each incoming observation $p_j$, the algorithm computes three state variables:

\begin{enumerate}
    \item The dynamic centroid $\mu_{\mathcal{C}}$, defined as the mean coordinate of all valid points currently comprising $\mathcal{C}$.
    \item The temporal gap $\Delta t = t_j - t_{j-1}$ between consecutive observations.
    \item The spatial displacement $d = \text{Haversine}(\mu_{\mathcal{C}}, p_j)$ separating the new observation from the centroid.
\end{enumerate}
The algorithm first evaluates temporal continuity. If the temporal gap $\Delta t$ exceeds the dropout tolerance $\tau_{max}$, the agent's behavior during the unobserved interval cannot be reliably inferred. The sequence expansion terminates immediately. 
Unlike the classic sliding-window method~\cite{li2008mining} used by Trackintel, SSPE does not close the window at the first distance violation; it first checks subsequent observations, and it measures distance from a dynamically updated centroid.

Provided the temporal link remains intact, the algorithm evaluates spatial cohesion. If $d \le \delta_{dist}$, the observation $p_j$ resides within the spatial neighborhood of the ongoing stay. It is appended to $\mathcal{C}$, the evaluation pointer advances ($j \leftarrow j + 1$), and the centroid $\mu_{\mathcal{C}}$ is updated in the subsequent iteration.

Conversely, if $d > \delta_{dist}$, the algorithm does not immediately confirm a physical departure. Instead, it activates the lookahead heuristic, scanning up to $n$ subsequent observations $p_k$, where $k \in [j+1, \min(j+n, N)]$. For each $p_k$, it calculates the future spatial displacement $d_{future} = \text{Haversine}(\mu_{\mathcal{C}}, p_k)$ and the accumulated temporal gap $\Delta t_{future} = t_k - t_{j-1}$. If any $p_k$ satisfies both $d_{future} \le \delta_{dist}$ and $\Delta t_{future} \le \tau_{max}$, the trajectory has rapidly returned to the stay region. The initial violation $p_j$ is classified as a transient positioning anomaly and bypassed without being added to $\mathcal{C}$, the expansion continues by incrementing $j \leftarrow j + 1$. If no such return occurs within the $n$-step window, the spatial deviation is confirmed as valid movement, and expansion terminates.

Upon termination of the candidate sequence, the total observed dwell duration $D = t_{\text{last}} - t_{\text{first}}$ is evaluated over the points in $\mathcal{C}$. Finally, the algorithm verifies the duration of the candidate sequence. If the duration meets the threshold $\tau_{min}$, the algorithm formalizes the staypoint $s = (uid, \mu_{\mathcal{C}}, t_{\text{first}}, t_{\text{last}})$, appends it to the output set $\mathcal{S}$, and updates the anchor pointer to bypass the completed sequence by setting $i \leftarrow \text{index}(p_{\text{last}}) + 1$. If the duration is less than $\tau_{min}$, the sequence is discarded as a transient pass-through, and the anchor increments by a single position with $i \leftarrow i + 1$.

\begin{algorithm}[t]
\caption{Sequential Staypoint Extraction (SSPE)}
\label{alg:sspe}
\begin{algorithmic}[1]

\fontsize{8pt}{11pt}\selectfont
\setlength{\baselineskip}{8.5pt}

\Statex \textbf{Input:} Trajectory $\mathcal{T} = \{p_1, p_2, \dots, p_N\}$, where $p = (uid, lat, lon, t)$
\Statex \textbf{Parameters:} $\delta_{dist}$ (Spatial threshold), $\tau_{min}$ (Min duration), $\tau_{max}$ (Max dropout), $n$ (Lookahead step)
\Statex \textbf{Output:} Set of Staypoints $\mathcal{S}$
\Statex
\State Initialize $\mathcal{S} \leftarrow \emptyset$, and pointer $i \leftarrow 1$
\While{$i < N$}
    \State Initialize candidate sequence $\mathcal{C} \leftarrow \{p_i\}$
    \State Set evaluation index $j \leftarrow i + 1$
    \While{$j \le N$}
        \State Compute dynamic centroid $\mu_{\mathcal{C}}$ as the mean coordinate of points in $\mathcal{C}$
        \State Calculate temporal gap $\Delta t \leftarrow t_j - t_{j-1}$
        \State Calculate spatial displacement $d \leftarrow \text{Haversine}(\mu_{\mathcal{C}}, p_j)$
        
        \If{$\Delta t > \tau_{max}$}
            \State \textbf{break} \Comment{Signal dropout exceeds tolerance}
        \EndIf
        
        \If{$d \le \delta_{dist}$}
            \State Append $p_j$ to $\mathcal{C}$
            \State $j \leftarrow j + 1$
        \Else \Comment{Distance threshold breached}
            \State Initialize $is\_noise \leftarrow \text{False}$
            \For{$k$ from $j+1$ to $\min(j+n, N)$} \Comment{Lookahead}
                \State Calculate $d_{future} \leftarrow \text{Haversine}(\mu_{\mathcal{C}}, p_k)$
                \State Calculate $\Delta t_{future} \leftarrow t_k - t_{j-1}$
                \If{$d_{future} \le \delta_{dist}$ and $\Delta t_{future} \le \tau_{max}$}
                    \State $is\_noise \leftarrow \text{True}$
                    \State \textbf{break}
                \EndIf
            \EndFor
            
            \If{$is\_noise$}
                \State $j \leftarrow j + 1$ \Comment{Classify $p_j$ as an anomaly and continue}
            \Else
                \State \textbf{break} \Comment{Spatial departure detected}
            \EndIf
        \EndIf
    \EndWhile
    
    \State Evaluate Duration $D \leftarrow t_{\text{last}} - t_{\text{first}}$ for $p \in \mathcal{C}$
    \If{$D \ge \tau_{min}$}
        \State Construct agent level staypoint $s = (uid, \mu_{\mathcal{C}}, t_{\text{first}}, t_{\text{last}} )$
        \State $\mathcal{S} \leftarrow \mathcal{S} \cup \{s\}$
        \State Update $i \leftarrow \text{index}(p_{\text{last}}) + 1$
    \Else
        \State $i \leftarrow i + 1$
    \EndIf
\EndWhile
\State \Return $\mathcal{S}$
\end{algorithmic}
\end{algorithm}
\subsection{Parameter Space Initialization}\label{sec:estimation}
As established, the efficacy of the SSPE algorithm depends on the appropriate configuration of its four parameters. However, real-world trajectory datasets typically lack the ground-truth annotations requisite for supervised parameter tuning. 
Relying on arbitrary default values or ad-hoc manual calibration across unknown or complex datasets frequently yields suboptimal extraction performance.

To address this challenge and enable robust deployment under completely unlabeled conditions, we propose Agreement-Based Parameter Estimation (ABPE).

\subsubsection{Agreement-Based Parameter Estimation}The parameter estimation framework conducts a joint search over a geometric grid bounded by the SSPE parameters. The boundaries of this grid are intrinsically derived from two primary data statistics: the characteristic sampling interval $g$ and the empirical GPS jitter scale $\sigma$. To ensure scalability without compromising representativeness, both the statistical derivation and the subsequent consensus search can be executed on a spatiotemporal subsample of the data. To construct this grid, we first define the spatial step distance $s$ and the temporal gap $\Delta t$ between consecutive observations for a given user $u$:
\begin{equation}s(u, i) = | p(u, i) - p(u, i-1) |\end{equation}
\begin{equation}\Delta t(u, i) = t(u, i) - t(u, i-1)\end{equation}

The baseline sampling interval $g$ is defined as the median of all strictly positive temporal gaps pooled across all users. To isolate pure spatial noise from actual travel, we filter the dataset to include only "normal" steps, which we define as those satisfying $\Delta t \le 5g$. 

Because step distances span multiple orders of magnitude, we apply a logarithmic transformation $z = \log_{10}(s+1)$, under which step distances form two modes: stationary noise and active movement. We partition these modes by applying Otsu's thresholding method \cite{otsu1979threshold} across a 200-bin histogram of $z$. 

For any candidate threshold cut $c$, the data is divided into a stationary class $C_0$ ($z \le c$) and a moving class $C_1$ ($z > c$). Let $w_0(c)$ and $w_1(c)$ denote the proportions of these classes, and let $\mu_0(c)$ and $\mu_1(c)$ denote their respective mean logarithmic distances. Otsu's method exhaustively searches for the optimal cut $c^*$ that maximizes the between-class variance:
\begin{equation}\sigma_B^2(c) = w_0(c) w_1(c) \big(\mu_0(c) - \mu_1(c)\big)^2\end{equation}
\begin{equation}c^* = \arg\max_{c} \sigma_B^2(c)\end{equation}

By identifying the cut $c$ that maximizes the between-class variance, we obtain a physical distance threshold:
\begin{equation}
\varepsilon = 10^{c^*} - 1
\end{equation}

Spatial steps satisfying $s \le \varepsilon$ are classified as stationary GPS noise. The global jitter scale $\sigma$ is defined as the 90th percentile of these stationary steps. Thus, 90\% of consecutive stationary step distances are at most $\sigma$.

To establish the dropout tolerance, we apply the same Otsu thresholding procedure to the log-transformed temporal intervals, yielding a temporal cut \(\tau _{Otsu}\). 
To prevent underestimating nominal sampling variance, the baseline dropout tolerance is lower-bounded by $5g$:
\begin{equation}
\tau_{base} = \max(\tau_{Otsu}, 5g)
\end{equation}

Next, we empirically estimate the baseline lookahead window $n$ by modeling the persistence of spatial anomalies. 
A spatial jump occurs at index $i$ if the step distance exceeds the noise threshold ($s(u, i) > \varepsilon$). 
For each jump, we evaluate up to a number of subsequent observations to determine if the trajectory reverts to the pre-jump location $p(u, i-1)$. 
Specifically, we find the minimal recovery length $L \in \{1, \dots, 5\}$ that satisfies both spatial and temporal constraints:
\begin{equation}| p(u, i+L) - p(u, i-1) | \le \varepsilon \end{equation}
\begin{equation}t(u, i+L) - t(u, i-1) \le \tau_{base}\end{equation}
If no such $L$ exists, the divergence is classified as a true physical departure. Otherwise, the event is recorded as a transient glitch of length $L$. 
The empirical lookahead parameter $n$ is defined as the ceiling of the 90th percentile of all recorded glitch lengths:\begin{equation}n = \lceil P_{90}(L) \rceil\end{equation}

With the metrics established, we construct the multi-dimensional search grid: 
The spatial threshold candidate set $\mathbf{\Delta}_{dist}$ is derived by scaling the jitter baseline $\sigma$. Depending on the computational budget, this is formulated either as a fine geometric progression:
\begin{equation}
\mathbf{\Delta}{dist} = { \sigma \cdot 1.25^k \mid k \in \{0, \dots, 7\} }
\end{equation}
or as a coarse progression utilizing predefined multipliers $\{1, 1.5, 2, 3, 4, 5\}$. 
Similarly, the maximum dropout search space $\tau_{max}$ scales geometrically from the expected sampling variance:
\begin{equation}
\tau_{max} = { 5g \cdot 2^k \mid k \in \{0, \dots, 8\} }
\end{equation}

In contrast, the minimum dwell duration $\tau_{min}$ is not estimated from the data distributions. Instead, it is defined strictly by the conceptual requirements of a staypoint (e.g., a 5-minute minimum dwell time) to filter out transient pass-throughs.

Given this defined parameter space, the following section details our consensus-based evaluation mechanism, which isolates the optimal configuration by maximizing spatiotemporal agreement across the extraction outputs.

\subsubsection{Consensus-Based Parameter Selection}
In the absence of ground truth, direct empirical validation of parameter configurations is infeasible. To address this, we formulate parameter selection as a consensus-maximization problem. We evaluate each candidate configuration by calculating the pairwise structural agreement of its extracted staypoints against all other configurations in the search space, selecting the parameter set that yields the highest global mean consensus.
\begin{table}[!htbp]
\centering
\newcommand{\best}[1]{\textbf{#1}}
\caption{F1, precision, and recall on the ExtraSensory dataset.}
\vspace{-.3cm}
\label{tab:extrasensory-results}
\begin{tabular}{lccc}
\toprule
\textbf{Method} & \textbf{F1} & \textbf{Precision} & \textbf{Recall} \\
\midrule
SSPE                & \best{0.7878} & \best{0.7226} & \best{0.8659} \\
Trackintel Baseline & 0.5098      & 0.4342      & 0.6173      \\
\midrule
SSPE (ABPE)  & \best{0.8107} & \best{0.8854} & \best{0.7476} \\
Trackintel (Hyperband) & 0.6305   & 0.5747        & 0.6983        \\
\bottomrule
\end{tabular}
\end{table}

For a given parameter configuration $\theta$, the SSPE algorithm generates a set of detected staypoints $D(\theta)$. 
Each stay $a \in D(\theta)$ is defined by a user identifier $u(a)$, a spatial centroid $(\phi(a), \lambda(a))$, and temporal boundaries $t_s(a)$ and $t_e(a)$. 
Two staypoints $x$ and $y$ are defined as a structural match, denoted $m(x, y)$, if and only if they satisfy fixed spatiotemporal tolerances:
\begin{equation}u(x) = u(y), \quad |\phi(x) - \phi(y)| \le r, \quad |\lambda(x) - \lambda(y)| \le r\end{equation}\begin{equation}|t_s(x) - t_s(y)| \le t, \quad |t_e(x) - t_e(y)| \le t\end{equation}

where $r$ is a fixed spatial tolerance (e.g., $0.001^{\circ}$) and $t$ is a fixed temporal tolerance (e.g., 5 minutes). 
These match the criteria used in our evaluation (Section \ref{exp}). For two detection sets $A$ and $B$, the directional coverage of $A$ by $B$, denoted $\text{cov}(A \mid B)$, is defined as the fraction of stays in $A$ possessing at least one match in $B$:
\begin{equation}\text{cov}(A \mid B) = \frac{|{ a \in A \mid \exists b \in B, m(a, b) }|}{|A|}\end{equation}
The pairwise structural agreement between sets $A$ and $B$ is quantified using the symmetric $F_1$ score, which computes the harmonic mean of their cross-coverages:
\begin{equation}F(A, B) = \frac{2 \cdot \text{cov}(A \mid B) \cdot \text{cov}(B \mid A)}{\text{cov}(A \mid B) + \text{cov}(B \mid A)}\end{equation}

To evaluate the overall grid, let $\Theta = \{\theta_1, \dots, \theta_n\}$ denote the candidate parameter space and $\mathcal{R} \subseteq \Theta$ a reference subset. 
The consensus scoring function is expressed as:

\begin{equation}C(\theta_i)=\frac{1}{|\mathcal R_i|}\sum_{j\in\mathcal R_i}F\big(D(\theta_i),D(\theta_j)\big)\end{equation}

where \(D(\theta_i) \text{ and } D(\theta_j)\) are the stay sets detected under \(\theta_i\) and \(\theta_j\). The selected configuration is $\hat{\theta} = \arg\max_i C(\theta_i)$.


\section{Experiments and Results}\label{exp}

To evaluate SSPE against real-world GPS noise, we utilize the ExtraSensory dataset~\cite{8090454}. Because the dataset lacks explicit staypoint ground truth, we infer stationary behavior from user-reported activity labels~\cite{lowens2023deepstay}. To address the ambiguity of raw labels and privacy-induced data gaps, the initial estimates underwent manual refinement based on historical trajectories and user velocity. This yields a finalized evaluation subset of 34 users, encompassing 59,187 observations and 358 verified staypoints sampled at approximately one-minute intervals.

With this empirical ground truth established, we evaluate extraction accuracy by formulating a strict spatiotemporal matching criterion.
A staypoint is parameterized by its two-dimensional spatial centroid and its temporal arrival and departure boundaries. A detected staypoint is classified as a valid match to a ground-truth staypoint if and only if the absolute deviations across all dimensions fall within predefined tolerances. Specifically, the spatial displacement in both longitude and latitude must be bounded by a threshold $\tau_s = 0.001^\circ$, equivalent to approximately 111 meters. Concurrently, the discrepancies in both arrival and departure times must not exceed a temporal threshold $\tau_t = 5$ minutes.

We compute the precision, recall, and $F_1$-score across SSPE and Trackintel methods, the results of which are detailed in Table~\ref{tab:extrasensory-results}. The empirical results confirm that SSPE substantially outperforms traditional threshold-based extraction methods. On the real-world ExtraSensory dataset, as detailed in Table \ref{tab:extrasensory-results}, SSPE applied in its fully unsupervised, single-shot configuration achieves an $F_1$-score of 0.7878, significantly outperforming the standard sliding-window baseline of 0.5098. This performance gain is driven by substantial improvements in both precision and recall. Furthermore, when integrated with the consensus-based parameter search, SSPE achieves SOTA extraction accuracy with an $F_1$-score of 0.8107. These findings validate that SSPE's dynamic centroid and lookahead effectively decouple transient coordinate drift from true spatial departures. Furthermore, the $n$-step lookahead recovery prevents the artificial fragmentation of stays during prolonged signal occlusions. These accuracy gains are achieved without computational overhead. Runtime analysis demonstrates that SSPE processes a single agent's trajectory in an average of 2.738 seconds, executing slightly faster than the deterministic Trackintel baseline of 2.909 seconds.

\section{Conclusion}
In this paper, we presented the Sequential Staypoint Extraction algorithm to address the fundamental fragility of traditional staypoint detection methods when processing sparse and noisy trajectory streams. By utilizing a dynamic spatial centroid alongside an $n$-step lookahead heuristic and explicit temporal bridging, SSPE natively mitigates the effects of extreme coordinate drift and protracted signal occlusions. Furthermore, to enable deployment in unlabeled environments, we introduced a fully unsupervised parameter estimation framework. This approach leverages intrinsic data statistics to identify optimal parameter configurations by maximizing structural consensus. Extensive evaluations on real-world ExtraSensory dataset confirm that SSPE yields state-of-the-art accuracy. It significantly improves precision, recall, and overall $F_1$-scores compared to established baselines under severe data degradation, all while maintaining strict computational efficiency.

\bibliographystyle{ACM-Reference-Format}
\bibliography{refs/main}

\end{document}